\documentclass[letterpaper,10pt,conference]{ieeeconf}

\IEEEoverridecommandlockouts

\usepackage[utf8]{inputenc} 
\usepackage[T1]{fontenc}    
\usepackage{hyperref}       
\usepackage{url}            
\usepackage{booktabs}       
\usepackage{amsfonts}       
\usepackage{nicefrac}       
\usepackage{microtype}      

\usepackage{amsmath}
\usepackage{amssymb}

\usepackage{dblfloatfix}
\usepackage{graphicx} 
\usepackage[font=footnotesize]{caption}
\usepackage{subcaption}
\usepackage{wrapfig}
\usepackage{color}

\usepackage{algorithm}
\usepackage{algorithmic}

\usepackage[export]{adjustbox}

\newcommand{\bs}{\mathbf{s}}
\newcommand{\ba}{\mathbf{a}}
\newcommand{\dt}{{\Delta}t}
\newcommand{\errval}{\mathcal{E}_{val}^{(H)}}
\newcommand{\bA}{\mathbf{A}}

\newcommand{\Drand}{\mathcal{D}_{\textsc{rand}}}
\newcommand{\Drl}{\mathcal{D}_{\textsc{rl}}}
\newcommand{\Dexpert}{\mathcal{D}^*}

\newcommand{\Sigmaphi}{\Sigma_{\pi_\phi}}

\newcommand{\specialcell}[2][c]{%
  \begin{tabular}[#1]{@{}c@{}}#2\end{tabular}}

\title{\LARGE \bf
Neural Network Dynamics\\for Model-Based Deep Reinforcement Learning\\with Model-Free Fine-Tuning}

\author{
    Anusha Nagabandi, Gregory Kahn, Ronald S. Fearing, Sergey Levine\\
    University of California, Berkeley
}

\begin{document}

\maketitle
\thispagestyle{empty}
\pagestyle{empty}

\begin{abstract}
Model-free deep reinforcement learning algorithms have been shown to be capable of learning a wide range of robotic skills, but typically require a very large number of samples to achieve good performance. Model-based algorithms, in principle, can provide for much more efficient learning, but have proven difficult to extend to expressive, high-capacity models such as deep neural networks. In this work, we demonstrate that medium-sized neural network models can in fact be combined with model predictive control (MPC) to achieve excellent sample complexity in a model-based reinforcement learning algorithm, producing stable and plausible gaits to accomplish various complex locomotion tasks. We also propose using deep neural network dynamics models to initialize a model-free learner, in order to combine the sample efficiency of model-based approaches with the high task-specific performance of model-free methods. We empirically demonstrate on MuJoCo locomotion tasks that our pure model-based approach trained on just random action data can follow arbitrary trajectories with excellent sample efficiency, and that our hybrid algorithm can accelerate model-free learning on high-speed benchmark tasks, achieving sample efficiency gains of $3-5\times$ on swimmer, cheetah, hopper, and ant agents. Videos can be found at \url{https://sites.google.com/view/mbmf}
\end{abstract}

\section{Introduction}

Model-free deep reinforcement learning algorithms have been shown to be capable of learning a wide range of tasks, ranging from playing video games from images~\cite{Mnih2013_NIPS,Oh2016_ICML} to learning complex locomotion skills~\cite{Schulman2015_ICML}. However, such methods suffer from very high sample complexity, often requiring millions of samples to achieve good performance~\cite{Schulman2015_ICML}. Model-based reinforcement learning algorithms are generally regarded as being more efficient~\cite{deisenroth2013survey}. However, to achieve good sample efficiency, these model-based algorithms have conventionally used either simple function approximators~\cite{lioutikov2014sample} or Bayesian models that resist overfitting~\cite{Deisenroth2011_ICML} in order to effectively learn the dynamics using few samples. This makes them difficult to apply to a wide range of complex, high-dimensional tasks. Although a number of prior works have attempted to mitigate these shortcomings by using large, expressive neural networks to model the complex dynamical systems typically used in deep reinforcement learning benchmarks~\cite{OpenAIgym,todorov2012mujoco}, such models often do not perform well~\cite{Gu2016_ICML} and have been limited to relatively simple, low-dimensional tasks~\cite{mishra2017prediction}.

\begin{figure}[t]
\centering
\includegraphics[width=0.75\columnwidth]{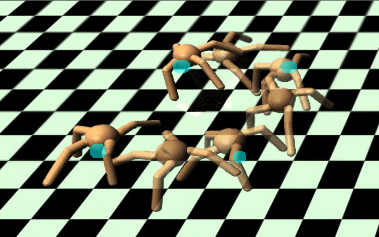}
\caption{Our method can learn a model that enables a simulated quadrupedal robot to autonomously discover a walking gait that follows user-defined waypoints at test time. Training for this task used $7\mathrm{e}5$ time steps, collected without any knowledge of the test-time navigation task.}
\label{fig:teaser}
\vspace*{-20pt}
\end{figure}

In this work, we demonstrate that multi-layer neural network models can in fact achieve excellent sample complexity in a model-based reinforcement learning algorithm, when combined with a few important design decisions such as data aggregation. The resulting models can then be used for model-based control, which we perform using model predictive control (MPC) with a simple random-sampling shooting method~\cite{Richards2004_MPC}. We demonstrate that this method can acquire effective locomotion gaits for a variety of MuJoCo benchmark systems, including the swimmer, half-cheetah, hopper, and ant. In fact, effective gaits can be obtained from models trained entirely off-policy, with data generated by taking only random actions. Fig.~\ref{fig:teaser} shows these models can be used at run-time to execute a variety of locomotion tasks such as trajectory following, where the agent can execute a path through a given set of sparse waypoints that represent desired center-of-mass positions. Additionally, less than four hours of random action data was needed for each system, indicating that the sample complexity of our model-based approach is low enough to be applied in the real world.

Although such model-based methods are drastically more sample efficient and more flexible than task-specific policies learned with model-free reinforcement learning, their asymptotic performance is usually worse than model-free learners due to model bias. Model-free algorithms are not limited by the accuracy of the model, and therefore can achieve better final performance, though at the expense of much higher sample complexity~\cite{deisenroth2013survey,kober2013reinforcement}. To address this issue, we use our model-based algorithm to initialize a model-free learner. The learned model-based controller provides good rollouts, which enable supervised initialization of a policy that can then be fine-tuned with model-free algorithms, such as policy gradients. We empirically demonstrate that the resulting hybrid model-based and model-free (Mb-Mf) algorithm can accelerate model-free learning, achieving sample efficiency gains of $3-5\times$ on the swimmer, cheetah, hopper, and ant MuJoCo locomotion benchmarks~\cite{todorov2012mujoco} as compared to pure model-free learning.

The primary contributions of our work are the following: (1) we demonstrate effective model-based reinforcement learning with neural network models for several contact-rich simulated locomotion tasks from standard deep reinforcement learning benchmarks, (2) we empirically evaluate a number of design decisions for neural network dynamics model learning, and (3) we show how a model-based learner can be used to initialize a model-free learner to achieve high rewards while drastically reducing sample complexity.

\section{Related Work}

Deep reinforcement learning algorithms based on Q-learning~\cite{mnih2015human,Oh2016_ICML,Gu2016_ICML}, actor-critic methods~\cite{Lillicrap2016_ICLR,Mnih2016_ICML,schulman2015high}, and policy gradients~\cite{Schulman2015_ICML,gu2016q} have been shown to learn very complex skills in high-dimensional state spaces, including simulated robotic locomotion, driving, video game playing, and navigation. However, the high sample complexity of purely model-free algorithms has made them difficult to use for learning in the real world, where sample collection is limited by the constraints of real-time operation. Model-based algorithms are known in general to outperform model-free learners in terms of sample complexity~\cite{deisenroth2013survey}, and in practice have been applied successfully to control robotic systems both in simulation and in the real world, such as pendulums~\cite{Deisenroth2011_ICML}, legged robots~\cite{morimoto2003minimax}, swimmers~\cite{meger2015learning}, and manipulators~\cite{deisenroth2011learning}. However, the most efficient model-based algorithms have used relatively simple function approximators, such as Gaussian processes~\cite{Deisenroth2011_ICML,boedecker2014approximate,ko2008gp}, time-varying linear models~\cite{lioutikov2014sample,levine2014learning,yip2014model}, and mixtures of Gaussians~\cite{khansari2011learning}. PILCO~\cite{Deisenroth2011_ICML}, in particular, is a model-based policy search method which reports excellent sample efficiency by learning probabilistic dynamics models and incorporating model uncertainty into long-term planning. These methods have difficulties, however, in high-dimensional spaces and with nonlinear dynamics. The most high-dimensional task demonstrated with PILCO that we could find has 11 dimensions~\cite{meger2015learning}, while the most complex task in our work has 49 dimensions and features challenging properties such as frictional contacts. To the best of our knowledge, no prior model-based method utilizing Gaussian processes has demonstrated successful learning for locomotion with frictional contacts, though several works have proposed to learn the dynamics, without demonstrating results on control~\cite{deisenroth2012toward}.

Although neural networks were widely used in earlier work to model plant dynamics ~\cite{hunt1992neural,bekey1992neural}, more recent model-based algorithms have achieved only limited success in applying such models to the more complex benchmark tasks that are commonly used in deep reinforcement learning. Several works have proposed to use deep neural network models for building predictive models of images~\cite{watter2015embed}, but these methods have either required extremely large datasets for training~\cite{watter2015embed} or were applied to short-horizon control tasks~\cite{wahlstrom2015pixels}. In contrast, we consider long-horizon simulated locomotion tasks, where the high-dimensional systems and contact-rich environment dynamics provide a considerable modeling challenge. \cite{mishra2017prediction} proposed a relatively complex time-convolutional model for dynamics prediction, but only demonstrated results on low-dimensional (2D) manipulation tasks. \cite{gal2016improving} extended PILCO~\cite{Deisenroth2011_ICML} using Bayesian neural networks, but only presented results on a low-dimensional cart-pole swingup task, which does not include frictional contacts.

Aside from training neural network dynamics models for model-based reinforcement learning, we also explore how such models can be used to accelerate a model-free learner. Prior work on model-based acceleration has explored a variety of avenues.  The classic Dyna~\cite{Sutton1991_AAAI} algorithm proposed to use a model to generate simulated experience that could be included in a model-free algorithm. This method was extended to work with deep neural network policies, but performed best with models that were not neural networks~\cite{Gu2016_ICML}. Other extensions to Dyna have also been proposed~\cite{silver2008sample,asadi2015strengths}. Model learning has also been used to accelerate model-free Bellman backups~\cite{heess2015learning}, but the gains in performance from including the model were relatively modest, compared to the $330\times, 26\times, 4\times,$ and $3\times$ speed-ups that we report from our hybrid Mb-Mf experiments. Prior work has also used model-based learners to guide policy optimization through supervised learning~\cite{levine2015end}, but the models that were used were typically local linear models. In a similar way, we also use supervised learning to initialize the policy, but we then fine-tune this policy with model-free learning to achieve the highest returns. Our model-based method is more flexible than local linear models, and it does not require multiple samples from the same initial state for local linearization.


\section{Preliminaries}
\label{sec:prelim}

The goal of reinforcement learning is to learn a policy that maximizes the sum of future rewards. At each time step $t$, the agent is in state $\bs_t \in \mathcal{S}$, executes some action $\ba_t \in \mathcal{A}$, receives reward $r_t = r(\bs_t, \ba_t)$, and transitions to the next state $\bs_{t+1}$ according to some unknown dynamics function $f : \mathcal{S} \times \mathcal{A} \rightarrow \mathcal{S}$. The goal at each time step is to take the action that maximizes the discounted sum of future rewards, given by $\sum_{t'=t}^\infty \gamma^{t'-t} r(\bs_{t'},\ba_{t'})$, where $\gamma \in [0,1]$ is a discount factor that prioritizes near-term rewards. Note that performing this policy extraction requires either knowing the underlying reward function $r(\bs_t, \ba_t)$ or estimating the reward function from samples~\cite{ng2000algorithms}. In this work, we assume access to the underlying reward function, which we use for planning actions under the learned model.

In model-based reinforcement learning, a model of the dynamics is used to make predictions, which is used for action selection. Let $\hat{f}_\theta(\bs_t,\ba_t)$ denote a learned discrete-time dynamics function, parameterized by $\theta$, that takes the current state $\bs_t$ and action $\ba_t$ and outputs an estimate of the next state at time $t + \dt$. We can then choose actions by solving the following optimization problem:
{\small
\begin{align}
(\ba_t, \dots, \ba_{t+H-1}) = \arg\max_{\ba_{t}, \dots, \ba_{t+H-1}} \sum_{t'=t}^{t+H-1} \gamma^{t'-t} r(\bs_{t'},\ba_{t'}) \label{eqn:policy-extraction}
\end{align}
}
\noindent 

In practice, it is often desirable to solve this optimization at each time step, execute only the first action $\ba_t$ from the sequence, and then replan at the next time step with updated state information. Such a control scheme is often referred to as model predictive control (MPC), and is known to compensate well for errors in the model.


\section{Model-Based Deep Reinforcement Learning}
\label{sec:mb}

We now present our model-based deep reinforcement learning algorithm. We detail our learned dynamics function $\hat{f}_\theta(\bs_t, \ba_t)$ in Sec.~\ref{sec:nn}, how to train the learned dynamics function in Sec.~\ref{sec:training}, how to extract a policy with our learned dynamics function in Sec.~\ref{sec:controller}, and how to use reinforcement learning to further improve our learned dynamics function in Sec.~\ref{sec:mb-rl}.


\subsection{Neural Network Dynamics Function}
\label{sec:nn}

We parameterize our learned dynamics function $\hat{f}_\theta(\bs_t, \ba_t)$ as a deep neural network, where the parameter vector $\theta$ represents the weights of the network. A straightforward parameterization for $\hat{f}_\theta(\bs_t, \ba_t)$ would take as input the current state $\bs_t$ and action $\ba_t$, and output the predicted next state $\hat{\bs}_{t+1}$. However, this function can be difficult to learn when the states $\bs_t$ and $\bs_{t+1}$ are too similar and the action has seemingly little effect on the output; this difficulty becomes more pronounced as the time between states $\dt$ becomes smaller and the state differences do not indicate the underlying dynamics well.

We overcome this issue by instead learning a dynamics function that predicts the change in state $\bs_t$ over the time step duration of $\dt$. Thus, the predicted next state is as follows: $\hat{\bs}_{t+1} = \bs_t + \hat{f}_\theta(\bs_t, \ba_t)$. Note that increasing this $\dt$ increases the information available from each data point, and can help with not only dynamics learning but also with planning using the learned dynamics model (Sec.~\ref{sec:controller}). However, increasing $\dt$ also increases the discretization and complexity of the underlying continuous-time dynamics, which can make the learning process more difficult.


\subsection{Training the Learned Dynamics Function}
\label{sec:training}

\textbf{Collecting training data}: We collect training data by sampling starting configurations  ${\bs_0 \sim p(\bs_0)}$, executing random actions at each timestep, and recording the resulting trajectories ${\tau = (\bs_0, \ba_0, \cdots, \bs_{T-2}, \ba_{T-2}, \bs_{T-1})}$ of length $T$. We note that these trajectories are very different from the trajectories the agents will end up executing when planning with this learned dynamics model and a given reward function $r(\bs_t, \ba_t)$ (Sec.~\ref{sec:controller}), showing the ability of model-based methods to learn from off-policy data.

\textbf{Data preprocessing}: We slice the trajectories $\{\tau\}$ into training data inputs $(\bs_t, \ba_t)$ and corresponding output labels $\bs_{t+1} - \bs_t$. We then subtract the mean of the data and divide by the standard deviation of the data to ensure the loss function weights the different parts of the state (e.g., positions and velocities) equally. We also add zero mean Gaussian noise to the training data (inputs and outputs) to increase model robustness. The training data is then stored in the dataset $\mathcal{D}$.

\textbf{Training the model}: We train the dynamics model $\hat{f}_\theta(\bs_t, \ba_t)$ by minimizing the error
\begin{align}
\mathcal{E}(\theta) = \frac{1}{|\mathcal{D}|} \sum_{(\bs_t, \ba_t, \bs_{t+1}) \in \mathcal{D}} \frac{1}{2} \| (\bs_{t+1} - \bs_t) - \hat{f}_\theta(\bs_t, \ba_t) \|^2 \label{eqn:our-mse}
\end{align}
using stochastic gradient descent. While training on the training dataset $\mathcal{D}$, we also calculate the mean squared error in Eqn.~\ref{eqn:our-mse} on a validation set $\mathcal{D}_{val}$, composed of trajectories not stored in the training dataset.

Although this error provides an estimate of how well our learned dynamics function is at predicting next state, we would in fact like to know how well our model can predict further into the future because we will ultimately use this model for longer-horizon control (Sec.~\ref{sec:controller}). We therefore calculate $H$-step validation errors by propagating the learned dynamics function forward $H$ times to make multi-step open-loop predictions. For each given sequence of true actions $(\ba_t, \dots \ba_{t+H-1})$ from $\mathcal{D}_{val}$, we compare the corresponding ground-truth states $(\bs_{t+1}, \dots \bs_{t+H})$ to the dynamics model's multi-step state predictions $(\hat{\bs}_{t+1}, \dots \hat{\bs}_{t+H})$, calculated as
\begin{align}
\errval &= \frac{1}{\mathcal{D}_{val}} \sum_{\mathcal{D}_{val}} \frac{1}{H} \sum_{h=1}^{H} \frac{1}{2} \| \bs_{t+h} - \hat{\bs}_{t+h} \|^2 ~~: \nonumber \\
\hat{\bs}_{t+h} &= \begin{cases}
    \bs_t & \quad h=0 \\
    \hat{\bs}_{t+h-1}+\hat{f}_\theta(\hat{\bs}_{t+h-1}, \ba_{t+h-1}) & \quad h>0
\end{cases}
\label{eqn:hstep-validation}
\end{align}%
This $H$-step validation is used to analyze our experimental results, but otherwise not used during training.


\subsection{Model-Based Control}
\label{sec:controller}

In order to use the learned model $\hat{f}_\theta(\bs_t, \ba_t)$, together with a reward function $r(\bs_t, \ba_t)$ that encodes some task, we formulate a model-based controller that is both computationally tractable and robust to inaccuracies in the learned dynamics model. Expanding on the discussion in Sec.~\ref{sec:prelim}, we first optimize the sequence of actions $\bA_t^{(H)} = (\ba_t, \cdots, \ba_{t+H-1})$ over a finite horizon $H$, using the learned dynamics model to predict future states:
\begin{align}
\bA_t^{(H)} = \arg\max_{\bA_t^{(H)}} &\sum_{t'=t}^{t+H-1} r(\hat{\bs}_{t'}, \ba_{t'}) ~~: \nonumber \\ 
\hat{\bs}_t &= \bs_t, \hat{\bs}_{t'+1} = \hat{\bs}_{t'} + \hat{f}_\theta(\hat{\bs}_{t'}, \ba_{t'}).
\label{eqn:controller-opt}
\end{align}

Calculating the exact optimum of Eqn.~\ref{eqn:controller-opt} is difficult due to the dynamics and reward functions being nonlinear, but many techniques exist for obtaining approximate solutions to finite-horizon control problems that are sufficient for succeeding at the desired task. In this work, we use a simple random-sampling shooting method~\cite{Rao2009_shooting} in which $K$ candidate action sequences are randomly generated, the corresponding state sequences are predicted using the learned dynamics model, the rewards for all sequences are calculated, and the candidate action sequence with the highest expected cumulative reward is chosen. Rather than have the policy execute this action sequence in open-loop, we use model predictive control (MPC): the policy executes only the first action $\ba_t$, receives updated state information $\bs_{t+1}$, and recalculates the optimal action sequence at the next time step. Note that for higher-dimensional action spaces and longer horizons, random sampling with MPC may be insufficient, and investigating other methods~\cite{Li2004_ilqr} in future work could improve performance.

Note that this combination of predictive dynamics model plus controller is beneficial in that the model is trained only once, but by simply changing the reward function, we can accomplish a variety of goals at run-time, without a need for live task-specific retraining.  


\subsection{Improving Model-Based Control with Reinforcement Learning}
\label{sec:mb-rl}

\begin{figure}[t]
  \centering
  \includegraphics[width=0.99\columnwidth]{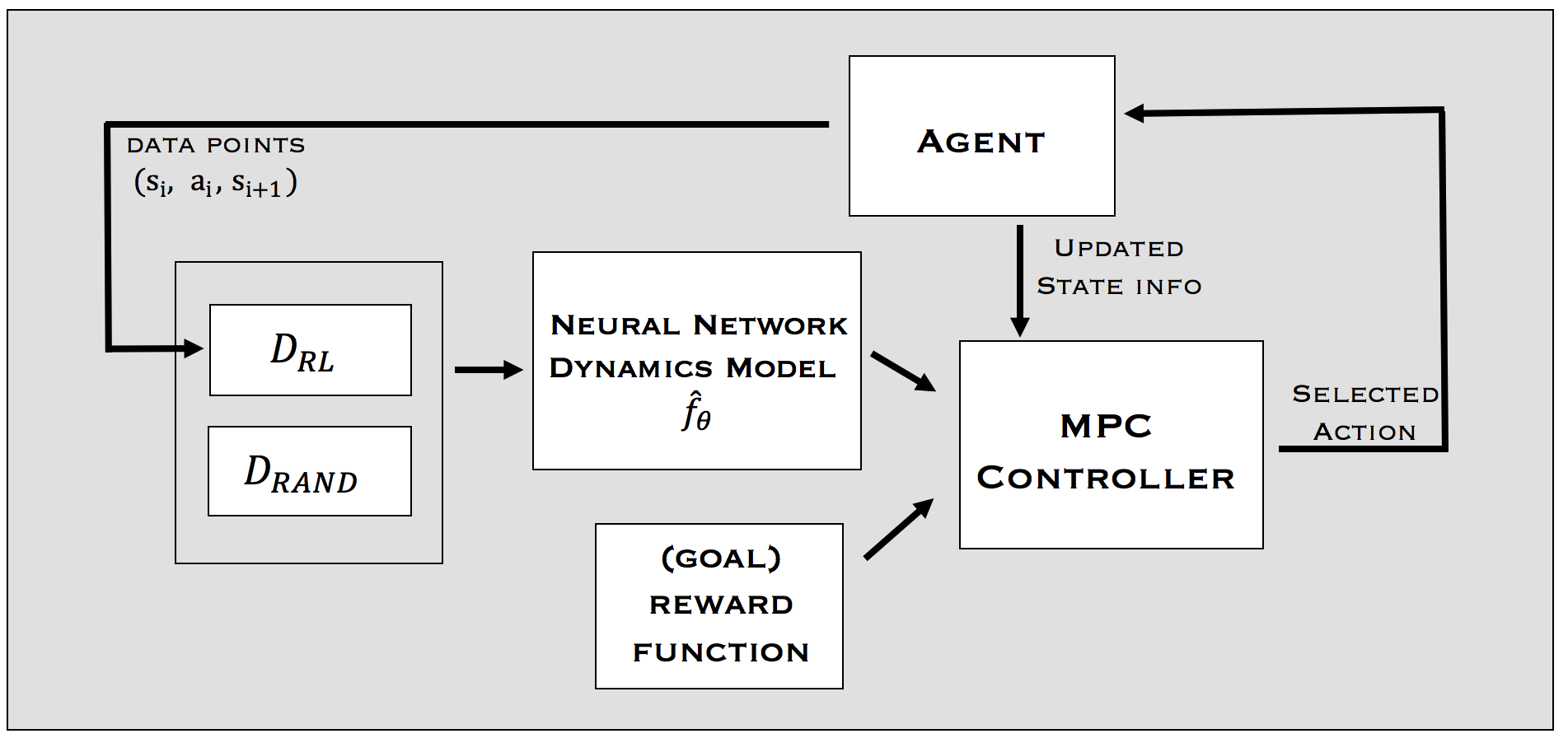}
  \caption{Illustration of Algorithm~\ref{alg:mb}. On the first iteration, random actions are performed and used to initialize $\Drand$. On all following iterations, this iterative procedure is used to train the dynamics model, run the MPC controller for action selection, aggregate data, and retrain the model.}
  \label{fig:block_diagram}
  \vspace*{-15pt}
\end{figure}

To improve the performance of our model-based learning algorithm, we gather additional on-policy data by alternating between gathering data with our current model and retraining our model using the aggregated data. This on-policy data aggregation (i.e., reinforcement learning) improves performance by mitigating the mismatch between the data's state-action distribution and the model-based controller's distribution~\cite{ross2011reduction}. Alg.~\ref{alg:mb} and Fig.~\ref{fig:block_diagram} provide an overview of our model-based reinforcement learning algorithm.

\begin{algorithm}[t]
  \caption{Model-based Reinforcement Learning}
  \label{alg:mb}
\begin{algorithmic}[1]
    \STATE gather dataset $\Drand$ of random trajectories
    \STATE initialize empty dataset $\Drl$, and randomly initialize $\hat{f}_\theta$
    \FOR{iter=1 {\bfseries to} max\_iter}
        \STATE train $\hat{f}_\theta(\bs, \ba)$ by performing gradient descent on Eqn.~\ref{eqn:our-mse}, using $\Drand$ and $\Drl$
        \FOR{$t=1$ {\bfseries to} $T$}
            \STATE get agent's current state $\bs_t$
            \STATE use $\hat{f}_\theta$ to estimate optimal action sequence $\bA^{(H)}_t$ (Eqn.~\ref{eqn:controller-opt})
            \STATE execute first action $\ba_t$ from selected action sequence $\bA^{(H)}_t$
            \STATE add $(\bs_t, \ba_t)$ to $\Drl$
        \ENDFOR
    \ENDFOR
\end{algorithmic}
\end{algorithm}

First, random trajectories are collected and added to dataset $\Drand$, which is used to train $\hat{f}_\theta$ by performing gradient descent on Eqn.~\ref{eqn:our-mse}. Then, the model-based MPC controller (Sec.~\ref{sec:controller}) gathers $T$ new on-policy datapoints and adds these datapoints to a separate dataset $\Drl$. The dynamics function $\hat{f}_\theta$ is then retrained using data from both $\Drand$ and $\Drl$. Note that during retraining, the neural network dynamics function's weights are warm-started with the weights from the previous iteration. The algorithm continues alternating between training the model and gathering additional data until a predefined maximum iteration is reached. We evaluate design decisions related to data aggregation in our experiments (Sec.~\ref{sec:ablation}).


\section{Mb-Mf: Model-Based Initialization of Model-Free Reinforcement Learning Algorithm}

The model-based reinforcement learning algorithm described above can learn complex gaits using very small numbers of samples, when compared to purely model-free learners. However, on benchmark tasks, its final performance still lags behind purely model-free algorithms. To achieve the best final results, we can combine the benefits of model-based and model-free learning by using the model-based learner to initialize a model-free learner. We propose a simple but highly effective method for combining our model-based approach with off-the-shelf, model-free methods by training a policy to mimic our learned model-based controller, and then using the resulting imitation policy as the initialization for a model-free reinforcement learning algorithm.

\subsection{Initializing the Model-Free Learner}

We first gather example trajectories with the MPC controller detailed in Sec.~\ref{sec:controller}, which uses the learned dynamics function $\hat{f}_\theta$ that was trained using our model-based reinforcement learning algorithm (Alg.~\ref{alg:mb}). We collect the trajectories into a dataset $\Dexpert$, and we then train a neural network policy $\pi_\phi(\ba | \bs)$ to match these ``expert'' trajectories in $\Dexpert$. We parameterize $\pi_\phi$ as a conditionally Gaussian policy $\pi_\phi(\ba | \bs) \sim \mathcal{N}(\mu_\phi(\bs), \Sigmaphi)$, in which the mean is parameterized by a neural network $\mu_\phi(\bs)$, and the covariance $\Sigmaphi$ is a fixed matrix. This policy's parameters are trained using the behavioral cloning objective
\begin{align}
\min_\phi \frac{1}{2} \sum_{(\bs_t, \ba_t) \in \Dexpert} ||\ba_t - \mu_\phi(\bs_t)||^2_2,
\label{eqn:imitation}
\end{align}
which we optimize using stochastic gradient descent. To achieve desired performance and address the data distribution problem, we applied DAGGER~\cite{ross2011reduction}: This consisted of iterations of training the policy, performing on-policy rollouts, querying the ``expert'' MPC controller for ``true'' action labels for those visited states, and then retraining the policy.

\begin{figure*}[!ht]
\centering
\begin{subfigure}{.24\textwidth}
  \centering
  \includegraphics[height=0.11\textheight]{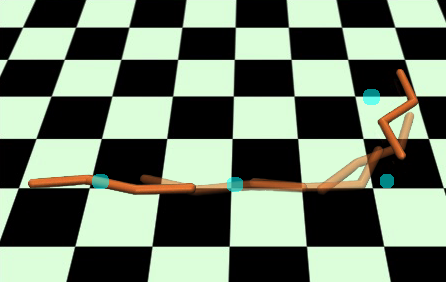}
  \caption{Swimmer left turn}
  \label{fig:traj-follow-swimmerleft}
\end{subfigure}
\hfill
\begin{subfigure}{.24\textwidth}
  \centering
  \includegraphics[height=0.11\textheight]{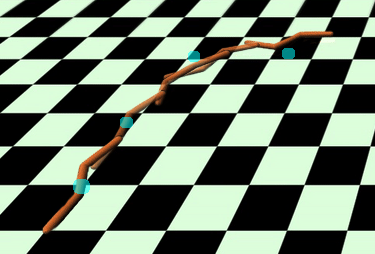}
  \caption{Swimmer right turn}
  \label{fig:traj-follow-swimmerright}
\end{subfigure}
\hfill
\begin{subfigure}{.24\textwidth}
  \centering
  \includegraphics[height=0.11\textheight]{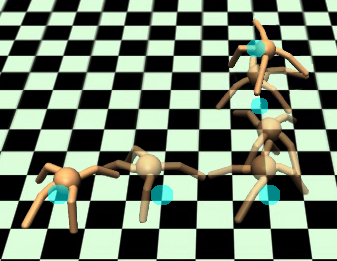}
  \caption{Ant left turn}
  \label{fig:traj-follow-antleft}
\end{subfigure}
\hfill
\begin{subfigure}{.24\textwidth}
  \centering
  \includegraphics[height=0.11\textheight]{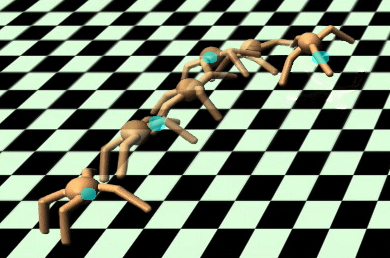}
  \caption{Ant right turn}
  \label{fig:traj-follow-antright}
\end{subfigure}
\caption{Trajectory following samples showing turns with swimmer and ant, with blue dots representing the center-of-mass positions that were specified as the desired trajectory. For each agent, we train the dynamics model only once on random trajectories, but use it at run-time to execute various desired trajectories.}
\label{fig:traj-follow}
  \vspace*{-15pt}
\end{figure*}

\subsection{Model-Free Reinforcement Learning}

After initialization, we can use the policy $\pi_\phi$, which was trained on data generated by our learned model-based controller, as an initial policy for a model-free reinforcement learning algorithm. Specifically, we use trust region policy optimization (TRPO)~\cite{Schulman2015_ICML}; such policy gradient algorithms are a good choice for model-free fine-tuning since they do not require any critic or value function for initialization~\cite{grondman2012survey}, though our method could also be combined with other model-free RL algorithms.

TRPO is also a common choice for the benchmark tasks we consider, and it provides us with a natural way to compare purely model-free learning with our model-based pre-initialization approach. Initializing TRPO with our learned expert policy $\pi_\phi$ is as simple as using $\pi_\phi$ as the initial policy for TRPO, instead of a standard randomly initialized policy. Although this approach of combining model-based and model-free methods is extremely simple, we demonstrate the efficacy of this approach in our experiments.

\section{Experimental Results}

We evaluated our model-based reinforcement learning approach (Alg.~\ref{alg:mb}) on agents in the MuJoCo~\cite{todorov2012mujoco} physics engine. The agents we used were swimmer ($\mathcal{S} \in \mathbb{R}^{16}, \mathcal{A} \in \mathbb{R}^{2}$), hopper ($\mathcal{S} \in \mathbb{R}^{17}, \mathcal{A} \in \mathbb{R}^{3}$), half-cheetah ($\mathcal{S} \in \mathbb{R}^{23}, \mathcal{A} \in \mathbb{R}^{6}$), and ant ($\mathcal{S} \in \mathbb{R}^{41}, \mathcal{A} \in \mathbb{R}^{8}$).  Relevant parameter values and implementation details are listed in the Appendix, and videos of all our experiments are provided online\footnote{\url{https://sites.google.com/view/mbmf}}.


\begin{figure}[!ht]
\centering
\begin{subfigure}{.24\columnwidth}
  \centering
  \includegraphics[height=0.055\textheight]{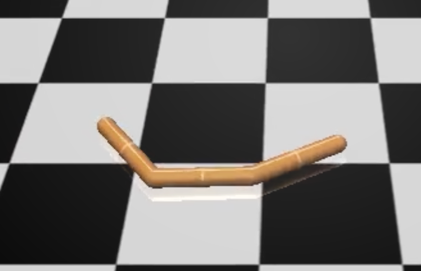}
  \caption{Swimmer}
\end{subfigure}
\begin{subfigure}{.24\columnwidth}
  \centering
  \includegraphics[height=0.055\textheight]{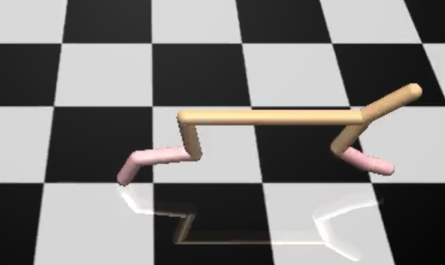}
  \caption{Cheetah}
\end{subfigure}
\begin{subfigure}{.24\columnwidth}
  \centering
  \includegraphics[height=0.055\textheight]{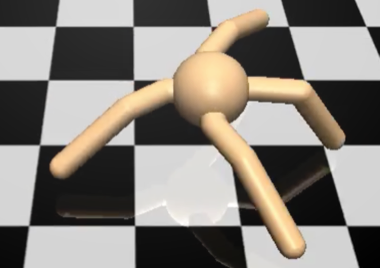}
  \caption{Ant}
\end{subfigure}
\begin{subfigure}{.16\columnwidth}
  \centering
  \includegraphics[height=0.055\textheight]{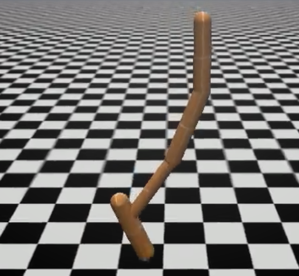}
  \caption{Hopper}
\end{subfigure}
\caption{Benchmark systems used in this paper. Agents on which we efficiently learn locomotion gaits, as well as combine our model-based approach with a model-free one to demonstrate fine-tuning performance.}
\label{fig:agents}
\vspace*{-15pt}
\end{figure}

\subsection{Evaluating Design Decisions for Model-Based Reinforcement Learning}
\label{sec:ablation}

We first evaluate various design decisions for model-based reinforcement learning with neural networks using empirical evaluations with our model-based approach (Sec.~\ref{sec:mb}). We explored these design decisions on the swimmer and half-cheetah agents on the locomotion task of running forward as quickly as possible. After each design decision was evaluated, we used the best outcome of that evaluation for the remainder of the evaluations.

\begin{figure*}[t]
\centering
    \includegraphics[width=0.7\linewidth]{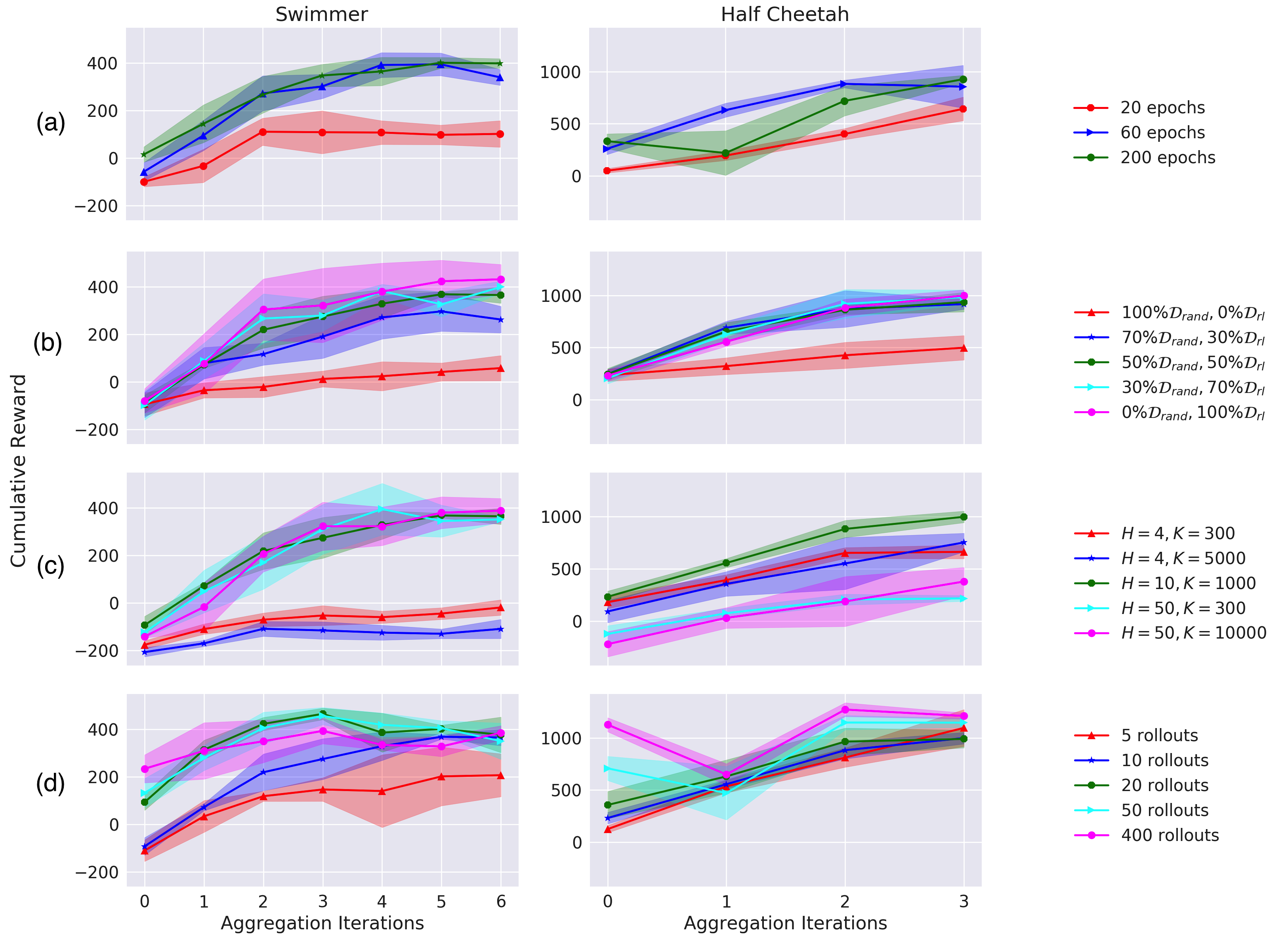}
\caption{Analysis of design decisions for our model-based reinforcement learning approach. (a) Training steps, 
(b) dataset training split, (c) horizon and number of actions sampled, (d) initial random trajectories. Training for more epochs, 
leveraging on-policy data, planning with medium-length horizons and many action samples were the best design choices, while data aggregation caused the number of initial trajectories that have little effect.}
\label{fig:ablations}
\vspace*{-20pt}
\end{figure*}

(A) Training steps. Fig.~\ref{fig:ablations}a shows varying numbers of gradient descent steps taken during the training of the learned dynamics function. As expected, training for too few epochs negatively affects learning performance, with 20 epochs causing swimmer to reach only half of the other experiments' performance.


(B) Dataset aggregation. Fig.~\ref{fig:ablations}b shows varying amounts of (initial) random data versus (aggregated) on-policy data used within each mini-batch of stochastic gradient descent when training the learned dynamics function. We see that training using mostly the aggregated on-policy rollouts significantly improves performance, revealing the benefits of improving learned models with reinforcement learning.

(C) Controller. Fig.~\ref{fig:ablations}c shows the effect of varying the horizon $H$ and the number of random samples $K$ used at each time step by the model-based controller. We see that too short of a horizon is harmful for performance, perhaps due to greedy behavior and entry into unrecoverable states. Additionally, the model-based controller for half-cheetah shows worse performance for longer horizons. This is further revealed below in Fig.~\ref{fig:forward_sim}, which illustrates a single 100-step validation rollout (as explained in Eqn.~\ref{eqn:hstep-validation}). We see here that the open-loop predictions for certain state elements, such as the center of mass x position, diverge from ground truth. Thus, a large $H$ leads to the use of an inaccurate model for making predictions, which is detrimental to task performance. Finally, with regards to the number of randomly sampled trajectories evaluated, we expect this value needing to be higher for systems with higher-dimensional action spaces.

(D) Number of initial random trajectories. Fig.~\ref{fig:ablations}d shows varying numbers of random trajectories used to initialize our model-based approach. We see that although a higher amount of initial training data leads to higher initial performance, data aggregation allows low-data initialization runs to reach a high final performance level, highlighting how on-policy data from reinforcement learning improves sample efficiency.

\begin{figure}[b]
  \vspace*{-15pt}
  \centering
  \includegraphics[width=0.95\columnwidth]{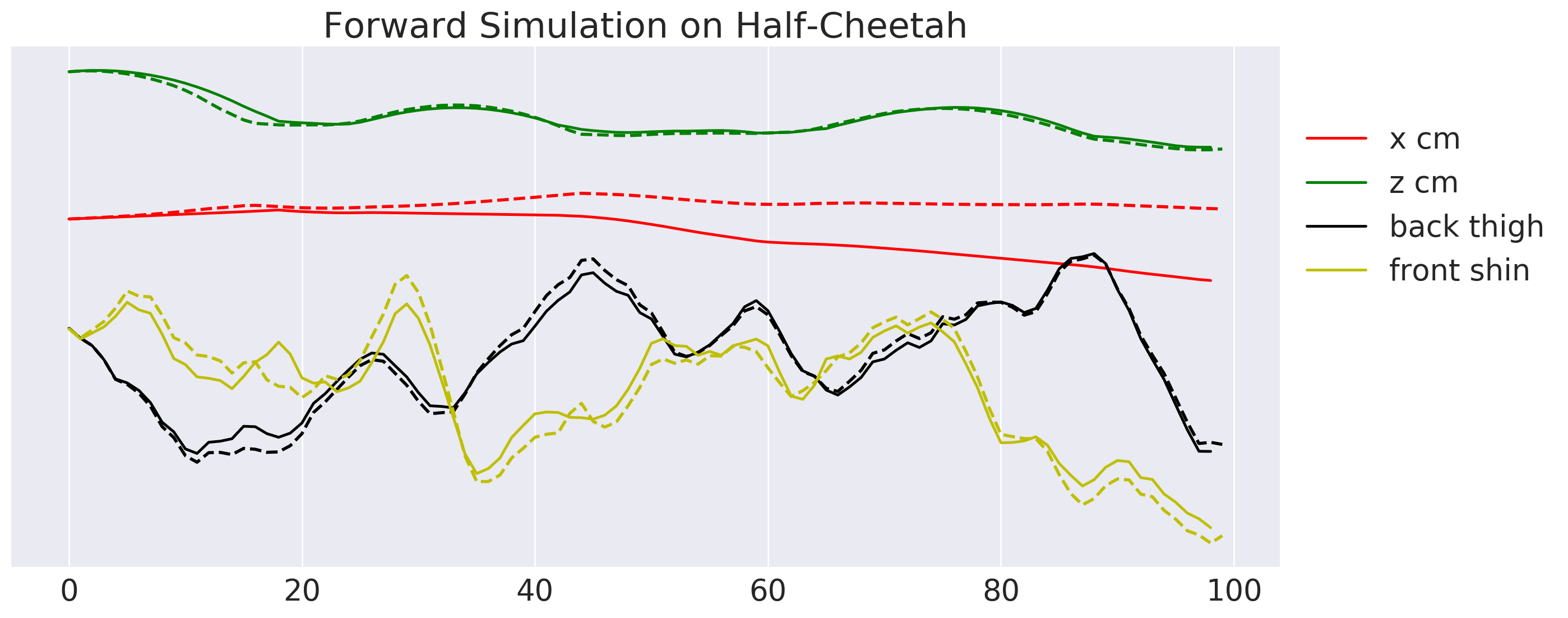}
  \caption{Given a fixed sequence of controls, we show the resulting true rollout (solid line) vs. the multi-step prediction from the learned dynamics model (dotted line) on the half-cheetah agent. Although we learn to predict certain elements of the state space well, note the eventual divergence of the learned model on some state elements when it is used to make multi-step open-loop predictions. However, our MPC-based controller with a short horizon can succeed in using the model to control an agent.}
\label{fig:forward_sim}
\end{figure}


\subsection{Trajectory Following with the Model-Based Controller}
\label{sec:traj_follow}

For the task of trajectory following, we evaluated our model-based reinforcement learning approach on the swimmer, ant, and half-cheetah environments (Fig.~\ref{fig:traj-follow}). Note that for these tasks, the dynamics model was trained using only random initial trajectories and was trained only once per agent, but the learned model was then used at run-time to accomplish different tasks. These results show that the models learned using our method are general enough to accommodate new tasks at test time, including tasks that are substantially more complex than anything that the robot did during training, such as following a curved path or making a U-turn. Furthermore, we show that even with the use of such a na\"{i}ve random-sampling controller, the learned dynamics model is powerful enough to perform a variety of tasks.

The reward function we use requires the robot to track the desired x/y center of mass positions. This reward consists of one term to penalize the perpendicular distance away from the desired trajectory, and a second term to encourage forward movement in the direction of the desired trajectory. The reward function does not tell the robot anything about how the limbs should be moved to accomplish the desired center of mass trajectory. The model-based algorithm must discover a suitable gait entirely on its own. Further details about this reward are included in the appendix.


\subsection{Mb-Mf Approach on Benchmark Tasks}
\label{sec:move_forward}

\begin{figure}[b]
  \centering
  \vspace*{-10pt}
  \includegraphics[width=0.8\linewidth]{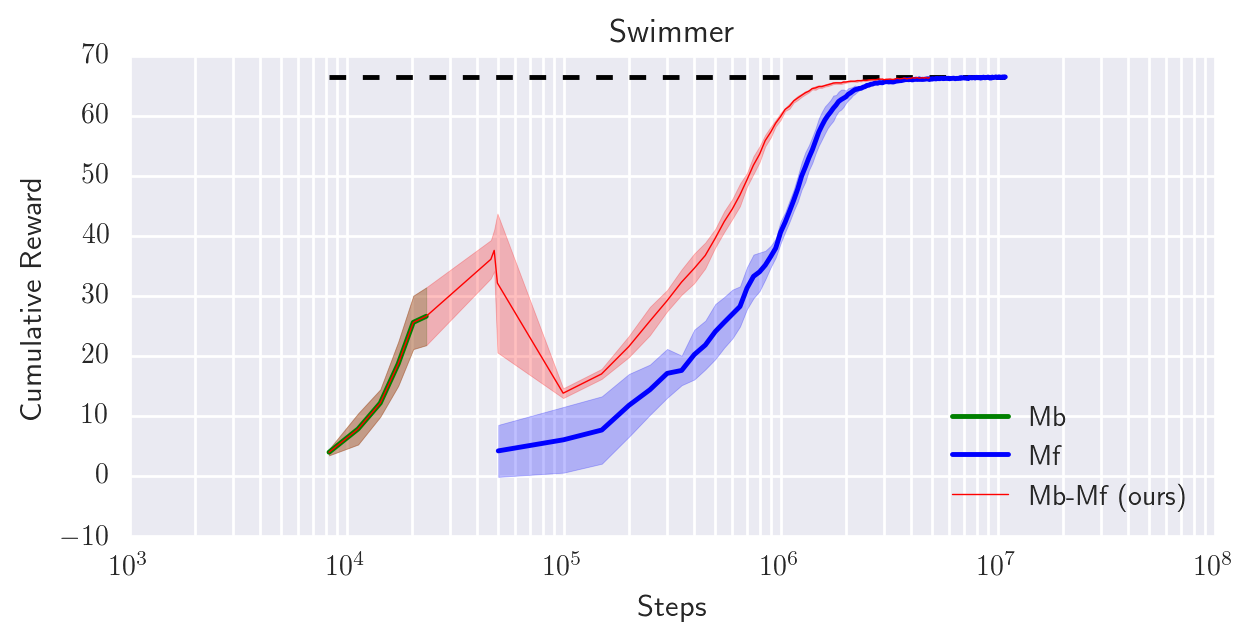}
  \caption{Using the standard Mujoco agent's reward function, our model-based method achieves a stable moving-forward gait for the swimmer using $20\times$ fewer data points than a model-free TRPO method. Furthermore, our hybrid Mb-Mf method allows TRPO to achieve its max performance $3\times$ faster than for TRPO alone.}
  \label{fig:mbmf-swimmer}
\end{figure}

\begin{figure*}
  \centering
  \includegraphics[width=0.32\textwidth]{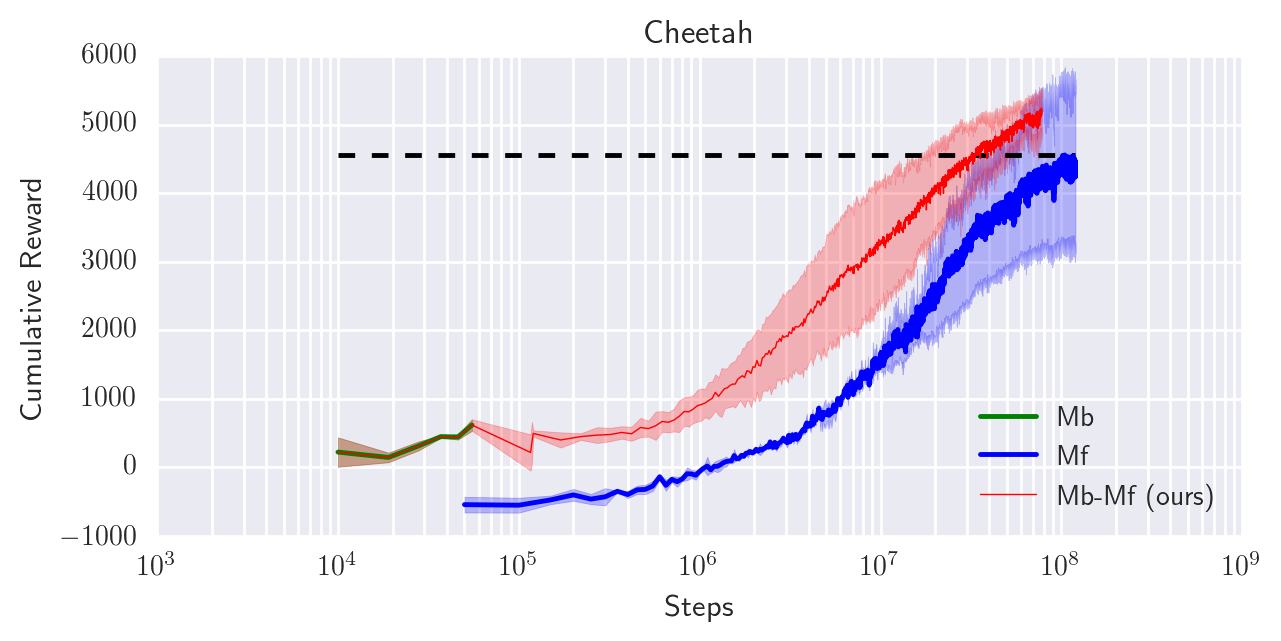}
  \includegraphics[width=0.32\textwidth]{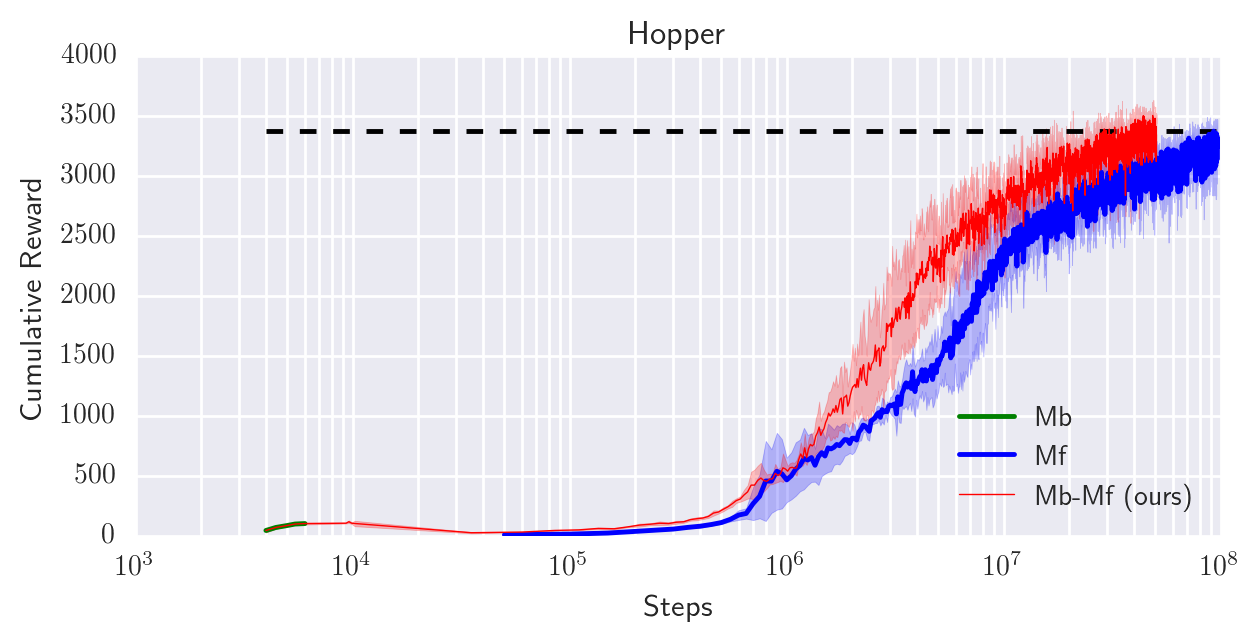}
  \includegraphics[width=0.32\textwidth]{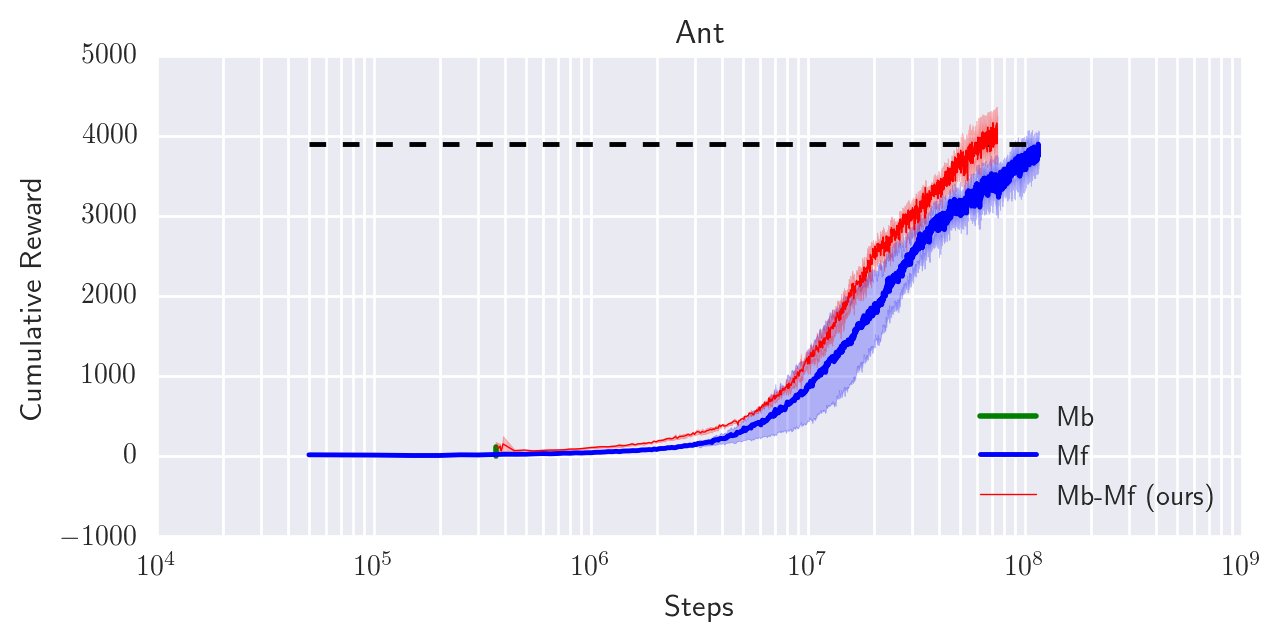}
  \caption{Plots show the mean and standard deviation over multiple runs and compare our model-based approach, a model-free approach (TRPO~\cite{Schulman2015_ICML}), and our hybrid model-based plus model-free approach. Our combined approach shows a $3-5\times$ improvement in sample efficiency for all shown agents. Note that the $x$-axis uses a logarithmic scale.}
  \label{fig:mbmf-results}
  \vspace*{-15pt}
\end{figure*}

We now compare our pure model-based approach with a pure model-free method on standard benchmark locomotion tasks, which require a simulated robot (swimmer, half-cheetah, hopper, or ant) to learn the fastest forward-moving gait possible. The model-free approach we compare with is the rllab~\cite{duan2016benchmarking} implementation of trust region policy optimization (TRPO)~\cite{Schulman2015_ICML}, which has obtained state-of-the-art results on these tasks.

For our model-based approach, we used the OpenAI gym~\cite{OpenAIgym} standard reward functions (listed in the appendix) for action selection in order to allow us to compare performance to model-free benchmarks. These reward functions primarily incentivize speed, and such high-level reward functions make it hard for our model-based method to succeed due to the myopic nature of the short-horizon MPC that we employ for action selection; therefore, the results of our model-based algorithm on all following plots are lower than would be if we designed our own reward function (for instance, a straight-line trajectory following reward function). 


Even with the extremely simplistic given reward functions, the agents can very quickly learn a gait that makes forward progress. The swimmer, for example, can quickly achieve qualitatively good moving forward behavior at $20\times$ faster than the model-free method. However, the final achieved rewards of our pure model-based approach were not sufficient to match the final performance of state-of-the-art model-free learners. Therefore, we combine our sample-efficient model-based method with a high-performing model-free method. In Fig.~\ref{fig:mbmf-results}, we show results comparing our pure model-based approach, a pure model-free approach (TRPO), and our hybrid Mb-Mf approach. 

With our pure model-based approach, these agents all learn a reasonable gait in very few steps. In the case of the hopper, our pure model-based approach learns to perform a double or triple hop very quickly in 1e4 steps, but performance plateaus as the reward signal of just forward velocity is not enough for the limited-horizon controller to keep the hopper upright for longer periods of time. Our hybrid Mb-Mf approach takes these quickly-learned gaits and performs model-free fine-tuning in order to achieve task success, achieving $3-5\times$ sample efficiency gains over pure model-free methods for all agents.

\section{Discussion}

We presented a model-based reinforcement learning algorithm that is able to learn neural network dynamics functions for complex simulated locomotion tasks using a small numbers of samples. Although a number of prior works have explored model-based learning with neural network dynamics models, our method achieves excellent performance on a number of challenging locomotion problems that exceed the complexity demonstrated in prior methods.

We described a number of important design decisions for effectively and efficiently training neural network dynamics models, and we presented detailed experiments that evaluated these design parameters. Our method quickly discovered a dynamics model that led to an effective gait, and that model could be applied to different trajectory following tasks at run-time, or the initial gait could then be fine-tuned with model-free learning to achieve high task rewards on benchmark Mujoco agents.


In addition to looking at the difference in sample complexity between our hybrid Mb-Mf approach and a pure model-free approach, there are also takeaways from the model-based approach alone. Our model-based algorithm cannot always reach extremely high rewards on its own, but it offers practical use by allowing the successful extraction of complex and realistic gaits. In general, our model-based approach can very quickly become competent at a task, whereas model-free approaches require immense amounts of data, but can become experts. For most practical applications, this competence that we achieve with our model-based approach is exactly what we need: For example, when we have a small legged robot with unknown dynamics and we want it to accomplish tasks in the real-world (such as exploration, construction, search and rescue, etc.), achieving reliable walking gaits that can follow any desired trajectory is a superior skill to that of just running straight forward as fast as possible. Additionally, consider the ant: A model-free approach requires 5e6 points to achieve a steady walking forward gait, but using just $14\%$ of those data points, our model-based approach can allow for travel in any direction and along arbitrary desired trajectories. Training such a dynamics model only once and applying it to various tasks is compelling; especially when looking toward application to real robots, this sample efficiency brings these methods out of the simulation world and into the realm of feasibility.

While the simplicity and effectiveness of our Mb-Mf approach is enticing for ease of practical application, an interesting avenue for future work is to integrate our model-based approach more tightly and elegantly with model-free learners (Q-learning, actor-critic methods), in order to provide further sample efficiency gains. 

Another exciting direction for future work is to deploy this method on real-world robotic systems, where the improved sample efficiency would make it practical to use even under the constraints of real-time sample collection in the real world. From the experiments shown in this paper, our method has shown applicability for systems with high-dimensional state spaces, systems with contact-rich environment dynamics, under-observed systems, and systems with complex nonlinear dynamics that provide a considerable modelling challenge. In addition to taking communication delays and computational limitations into account, another line of future work includes improving the MPC controller. In this paper, we chose to use a na\"{i}ve random-sampling controller to further emphasize the power of the learned dynamics models; however, this may not be feasible on real systems with limited computational power, or on systems with high-dimensional actions spaces that would require a large number of actions to sampled. Thus, further development of a real-time controller via optimization techniques is compelling future work.


\vspace*{-5pt}
\section{Acknowledgements}

We thank Vitchyr Pong for running some baseline experiments. This work is supported by the National Science Foundation under the National Robotics Initiative, Award CMMI-1427096, as well as IIS-1637443, IIS-1614653, and a NSF Graduate Research Fellowship.

\vspace*{-5pt}
\bibliographystyle{IEEEtran}
\bibliography{2018_icra_mime}
\vspace*{-5pt}

\clearpage
\appendix
\subsection{Experimental Details for Model-Based approach}

In the tables of parameters listed below, F represents the task of moving forward and TF represents the task of trajectory following. In our implementation, we normalize all environments such that all actions fall in the range $[-\mathbf{1},\mathbf{1}]$.\\

\subsubsection{Collecting initial dataset $\Drand$}
We populated this initial dataset with rollouts that resulted from the execution of random actions $\ba_t \sim \textnormal{Uniform}[-\mathbf{1},\mathbf{1}$]. Each rollout started at some initial state $\bs_0$, but to help with further exploration of the state space, we added noise to this starting state. For all agents, we added noise $\sim \mathcal{N}(0,0.001)$ to the $\bs_0^{\textsc{pos}}$ and $\bs_0^{\textsc{vel}}$ elements of the state. The only exception to this starting state noise was for the swimmer on the task of trajectory following. To allow enough exploration of the state space to be able to execute arbitrary trajectories in the future, we found that we had to add more noise to the ``heading" element of the swimmer state: We swept this value across the full range of possible headings by adding noise $\sim \textnormal{Uniform}(-\pi, \pi)$. Below are the number of rollouts and length of each rollout we used for each domain and task:

\renewcommand{\arraystretch}{1.2}

\begin{table}[!ht]
\begin{tabular}{|l|c|c|c|c|}
\hline  & Swimmer TF & Half-Cheetah TF & Ant TF\\
\hline Number of rollouts &200 &200 &700\\
\hline Length of each rollout &500 &1000 &1000\\
\hline
\end{tabular}
\end{table}

\vspace*{-10pt}
\begin{table}[!ht]{
    \scriptsize
    \begin{tabular}{|l|c|c|c|c|}
        \hline  & Swimmer F & Half-Cheetah F &Hopper F & Ant F\\
        \hline Number of rollouts &25 &10 &20 &700\\
        \hline Length of each rollout &333 &1000 &200 &1000\\
        \hline
    \end{tabular}
}
\end{table}

\subsubsection{Training the dynamics function}
For all agents, the neural network architecture for our dynamics function has two hidden layers, each of dimension 500, with ReLU activations. We trained this dynamics function using the Adam optimizer~\cite{Kingma2014_ICLR} with learning rate 0.001 and batch size 512. Prior to training, both the inputs and outputs in the dataset were pre-processed to have mean $0$ and standard deviation $1$. Below are relevant parameters for our data aggregation procedure (Sec.~\ref{sec:mb-rl}):

\renewcommand{\arraystretch}{1.2}

\begin{table}[!ht]
\begin{tabular}{|l|c|c|c|c|}
\hline  & Swimmer TF & Half-Cheetah TF & Ant TF\\
\hline Training epochs &70 &40 &60\\
\hline Aggregation iters &0 &0 &0\\
\hline
\end{tabular}
\end{table}

\vspace*{-10pt}
\begin{table}[!ht]
{\scriptsize
\begin{tabular}{|l|c|c|c|c|}
\hline  & Swimmer F & Half-Cheetah F & Hopper F & Ant F\\
\hline \specialcell{Training epochs\\per aggregation iter} &30 &60 &40 &20\\
\hline Aggregation iters &6 &7 &5 &*N/A\\
\hline \specialcell{Rollouts added\\per aggregation iter} &9 &9 &10 &*N/A\\
\hline \specialcell{Length of\\aggregated rollouts} &333 &1000 &**N/A &*N/A\\
\hline \specialcell{$\Drand$-$\Drl$ split\\for retraining} &10-90 &10-90 &10-90 &*N/A\\
\hline
\end{tabular}
}
\end{table}
{\footnotesize
\noindent
*N/A because no aggregation performed\\
**N/A because each rollout was of different length, depending on termination conditions of that agent
}

\newpage
\subsubsection{Other}
\renewcommand{\arraystretch}{1.2}

Additional model-based hyperparameters

\begin{table}[h]
{\footnotesize
\begin{tabular}{|l|c|c|c|c|}
\hline  & Swimmer TF & Half-Cheetah TF & Ant TF\\
\hline Timestep $dt$ &0.15s &0.01s &0.02s\\
\hline Controller horizon $H$ &5 &10 &15\\
\hline \specialcell{Number actions\\sampled $K$} &5000 &1000 &7000\\
\hline
\end{tabular}
}
\end{table}

\vspace*{-5pt}
\begin{table}[h]
{\footnotesize
\begin{tabular}{|l|c|c|c|c|}
\hline  & Swimmer F & Half-Cheetah F & Hopper F & Ant F\\
\hline Timestep $dt$ &0.15s &0.01s &0.02s &0.02s\\
\hline \specialcell{Controller\\horizon $H$} &20 &20 &40 &5\\
\hline \specialcell{Num. actions\\sampled $K$} &5000 &1000 &1000 &15000\\
\hline
\end{tabular}
\vspace*{-10pt}
}
\end{table}


\subsection{Experimental Details for Hybrid Mb-Mf approach}

For the task of moving forward, we first we saved rollouts from the execution of our MPC controller. At each time step during the collection of these rollouts, noise $\sim \mathcal{N}(0,0.005)$ was added to the optimal action before execution in order to promote exploration while still achieving good behavior. We then trained a Gaussian neural network policy to imitate these saved rollouts. This policy was then used to initialize the model-free TRPO algorithm. \\

\subsubsection{Imitation Learning}
We represented the mean of this policy as a neural network composed of tanh nonlinearities and two hidden layers, each of dimension 64. We trained this policy using the Adam optimizer~\cite{Kingma2014_ICLR} with learning rate 0.0001 and batchsize 500. In addition to training the mean network, the standard deviation (std) was another parameter of importance. Optimizing this std parameter according to the imitation learning loss function resulted in worse TRPO performance than arbitrarily using a larger std, perhaps because a higher std on the initial policy leads to more exploration and thus is more beneficial to TRPO. Therefore, we trained our policy's mean network using the standard imitation learning loss function, but we manually selected the std to be 1.0.
\renewcommand{\arraystretch}{1.2}
\begin{table}[!ht]
\begin{tabular}{|l|c|c|c|c|}
\hline & Swimmer & Half-Cheetah & Hopper & Ant\\
\hline \specialcell{Number of saved\\MPC rollouts} &30 &30 &60 &30\\
\hline \specialcell{Avg rewards of saved\\MPC rollouts} &30 &600 &100 &110\\
\hline Number of DAGGER iters &3 &3 &5 &5\\
\hline \specialcell{Training epochs\\per DAGGER iter} &70 &300 &200 &200\\
\hline \specialcell{Rollouts aggregated\\per DAGGER iter} &5 &2 &5 &5\\
\hline \specialcell{Avg rewards of resulting\\imitation policy} &40 &500 &110 &150\\
\hline
\end{tabular}
\end{table}

\subsubsection{TRPO}
We used rllab's~\cite{duan2016benchmarking} implementation of the TRPO algorithm with the following parameters for all agents: batch size 50000, base eps $1\mathrm{e}-5$, discount factor 0.995, and step size 0.5.

\subsection{Reward Functions}

As described in the paper, we are tasked with performing action selection at each time step. To assign a notion of value to a given sequence of actions, we use the reward functions shown below.\\

\subsubsection{Trajectory Following}

We formulated a reward function to allow agents to follow trajectories, where the desired trajectory is specified as sparse and lower-dimensional guidance in the form of desired $(x,y)$ center of mass positions. 

We first convert the set of desired waypoints into a set of line segments for the agent to travel along. The reward function shown in Alg.~\ref{alg:traj_follow} computes the reward value $R$ of the given action sequence $\bA$, and it does so by penalizing perpendicular distance away from the desired trajectory while encouraging forward progress along the trajectory. 

This is only an example reward function, so the user can choose to improve task performance in their own implementation by adding other factors to the reward function, such as penalizing for jumping too high or for falling down. We note that the standard Mujoco reward functions have been similarly tuned to include components such as terminal conditions (e.g., agent falling).

\begin{algorithm}[h]
  \caption{Reward function for Trajectory Following}
  \label{alg:traj_follow}
\begin{algorithmic}[1]
    \STATE \textbf{input}: current true state $\bs_t$, \\ \hspace*{27pt} sequence of actions $\bA$=$\{\ba_0, \ba_1, \dots, \ba_{H-1}\}$, \\ \hspace*{27pt} set of desired line segments to follow $L$=$\{L_0, \dots, L_{x}\}$
    \STATE reward $R \leftarrow 0$
    \FOR{each action $\ba_t$ in $\bA$}
        \STATE get predicted next state $\hat{\bs}_{t+1} = \hat{f}_\theta(\hat{\bs}_{t}, \ba_t)$
        \STATE $L_c \leftarrow$ closest line segment in $L$ to the point $(\hat{\bs}_{t+1}^{\textsc{x}},\hat{\bs}_{t+1}^{\textsc{y}})$
        \STATE $\textnormal{proj}_t^\parallel, \textnormal{proj}_t^\bot \leftarrow$ project point $(\hat{\bs}_{t+1}^{\textsc{x}},\hat{\bs}_{t+1}^{\textsc{y}})$ onto $L_c$
        \STATE $R \leftarrow R -\alpha(\textnormal{proj}_t^{\bot}) +\beta(\textnormal{proj}_t^{\parallel}-\textnormal{proj}_{t-1}^{\parallel}) $
    \ENDFOR
    \STATE \textbf{return:} reward $R$
\end{algorithmic}
\end{algorithm}

\subsubsection{Moving Forward}
We list below the standard reward functions $r_t(\bs_t, \ba_t)$ for moving forward with Mujoco agents. As before, the reward $R$ corresponding to a given action sequence $\bA$ is calculated as $R = \sum\limits_{t=0}^{H-1} r_t $.

\renewcommand{\arraystretch}{2.2}
\begin{table}[h]
\centering
\begin{tabular}{|c|c|c|}
\hline  & Reward $r_t$\\
\hline Swimmer & $\bs_{t+1}^{\textsc{xvel}} - 0.5\|\frac{\ba}{50}\|_2^2$\\
\hline Half-Cheetah & $\bs_{t+1}^{\textsc{xvel}} - 0.05\|\frac{\ba}{1}\|_2^2$\\
\hline Hopper & $\bs_{t+1}^{\textsc{xvel}} + 1 - 0.005\|\frac{\ba}{200}\|_2^2$\\
\hline Ant & $\bs_{t+1}^{\textsc{xvel}} + 0.5 - 0.005\|\frac{\ba}{150}\|_2^2$\\
\hline
\end{tabular}
\end{table}

\end{document}